# DETECTING SMALL OBJECTS IN THERMAL IMAGES USING SINGLE-SHOT DETECTOR

**Hao Zhang**,

**Xiang-gong Hong**[*],

**Li Zhu**,

*School of Information Engineering, Nanchang University, Nanchang 330031, China*

E-mail: [*]maxh@email.ncu.edu.cn

**Abstract**–SSD (Single Shot Multibox Detector) is one of the most successful object detectors for its high accuracy and fast speed. However, the features from shallow layer (mainly Conv4_3) of SSD lack semantic information, resulting in poor performance in small objects. In this paper, we proposed DDSSD (Dilation and Deconvolution Single Shot Multibox Detector), an enhanced SSD with a novel feature fusion module which can improve the performance over SSD for small object detection. In the feature fusion module, dilation convolution module is utilized to enlarge the receptive field of features from shallow layer and deconvolution module is adopted to increase the size of feature maps from high layer. Our network achieves **79.7%** mAP on PASCAL VOC2007 test and **28.3%** mmAP on MS COCO test-dev at 41 FPS with only 300×300 input using a single Nvidia 1080 GPU. Especially, for small objects, DDSSD achieves **10.5%** on MS COCO and **22.8%** on FLIR thermal dataset, outperforming a lot of state-of-the-art object detection algorithms in both aspects of accuracy and speed.

*Keywords:* Dilation and Deconvolution Module; Object Detection; Small Objects; SSD; Thermal Image.

## 1. INTRODUCTION

Object detection is a fundamental and challenging task in computer vision. Deep learning approaches have shown impressive improvements in general object detection. However, there are still challenges to be addressed for objects of different scales. Detecting small objects, for example those in thermal images [1], is a quite challenging task due to their limited resolution and information in images.

Object detectors in deep learning era can be divided into two classes: one-stage and two-stage. In two-stage methods, a set of region proposals is first generated by Selective Search [2] or Edge boxes [3] or region proposal network (RPN) [4], then the accurate object locations and the corresponding category labels are determined by deep ConvNets. The selected regions ensure two-stage detectors perform excellent in accuracy perspective, while it is hard for two-stage approaches to detect images fast. By contrast, the one-stage approaches turn the prediction of the bounding boxes into a classification problem, which makes predictions in one time, and thus perform better in inference speed aspect. For example, SSD

[5] can detect one image in about 1.7 ms with better performance than Faster R-CNN [4] while running about 7 times faster. So, for fast small object detection, we only consider the one-stage object detectors for their well balance of accuracy-vs-speed. The Single Shot MultiBox Detector (SSD) takes the reduced VGG-16 [6] (16-layer version of Visual Geometry Group Nets) as based network and add extra convolution layers to the end of it. Multi-scale feature maps are utilized to detect objects with different scales within an image. Shallower layers are used to predict smaller objects, while deeper layers for bigger objects. However, shallower layers often lack in semantic information, which is important supplement for small object detection. Therefore, introducing the semantic information captured in convolutional forward computation back to the shallower layers will improve the detection performance of small objects. Additionally, as studied in [7], the receptive field of the prediction layers is not large enough for the size of the grid. Fig.1 shows the visualization of the feature maps from L2 Norm (an operation after Conv4_3) of the conventional SSD and from Dilation Module of our DDSSD, respectively. Obviously, under the same size (38×38), in the original SSD only several pixels are activated, which makes the detection difficult. By contrast, our DDSSD achieves a wider range of activation, which helps a lot when detecting especially for small objects. Unfortunately, dilation convolution also introduces additional noise as shown in the feature maps.

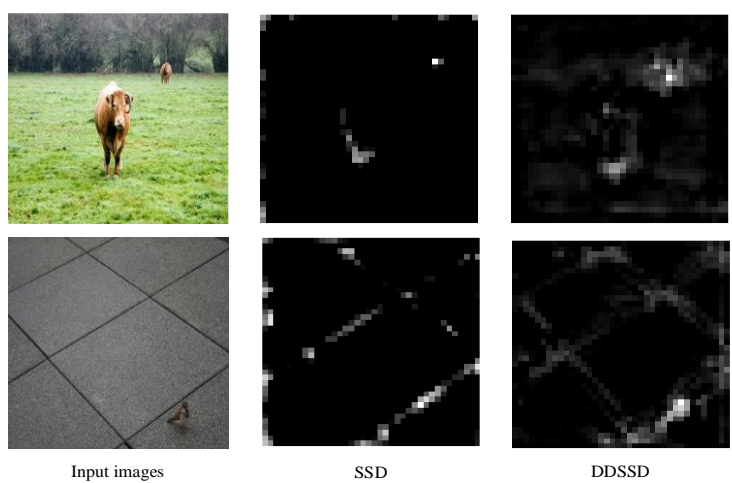

Input images     SSD     DDSSD

**Figure 1**. Visualization of feature maps.
The feature maps from SSD [5] are from the Conv4_3 after the L2 Norm operation, while DDSSD's are from Dilation Module which are utilized to expand the receptive field of the original SSD.
Note that all feature maps have a resolution of 38×38.

Based on the analysis above, we take the one-stage object detector SSD as our baseline with respect to accuracy-vs-speed trade-off. Specifically, Dilation Module is used to enlarge the receptive field in the VGG-16 feature extraction stage and then multi-scale feature fusion through Deconvolution Module is utilized to introduce contextual information for small object detection. The main contributions of this paper are summarized as:

1) Dilation and Deconvolution Module are proposed to enlarge the receptive field and introduce semantic information for small object detection.

2) Experimental results demonstrate the superiority of our algorithm compared to state-of-the-art algorithms. With the input size 300 × 300, DDSSD achieves 79.7% mAP on PASCAL VOC 2007 test, which is 2.2 and 1.1 points better than the conventional SSD and DSSD respectively. In addition, DDSSD performs well on MS COCO with a mmAP of 28.3% and 10.5% for small objects, which is comparable with many state-of-the-art methods. For extreme small objects in FLIR ADAS dataset, DDSSD achieves an mmAP of 22.8% and 49.9% for AP under IoU of 0.5.

## 2. RELATED WORK

Object detection: Because of the feature extraction power of deep convolution neural networks (DCNNs), object detectors such as OverFeat [8] and R-CNN [9] have begun to show the impressive improvements in accuracy in generic object detection. OverFeat [8] is the first to apply a CNN as a feature extractor in the sliding window on an image pyramid. R-CNN [9] adopts the region proposals generated from selective search [2] or Edge boxes [3] to generate the region-based feature from a pre-trained CNN and SVMs to do classification. SPPNet [10] utilizes a spatial pyramid pooling layer which allows the classification module to reuse the CNN feature regardless of the input image resolutions. Fast R-CNN [11] introduces multi-task loss to train the CNN with both the classification and location regression loss end to end. Faster R-CNN [4] suggests replacing selective search with a region proposal network (RPN). RPN is used to generate the candidate bounding boxes (anchor boxes) and filter out the background regions at the same time. Then another small network is used to do classification and bounding box location regression based on these proposals. R-FCN [12] replaces RoI (Region of Interests) pooling in Faster R-CNN with a position sensitive RoI pooling (PSRoI) to improve the detector's performance both on speed and accuracy aspect. Except for the two-stage detectors, there are also some efficient one-stage object detectors. YOLO (you only look once) [13] divides the input image into several grids and performs localization and classification on each part of the image. Due to this method, YOLO can run object detection at a very high speed but the accuracy is not satisfactory enough. YOLOv2 [14] is an enhanced version of YOLO and it improves the YOLO by removing the fully connected layers and adopts anchor boxes like the RPN.

**Multi-scale representation**: Multi-scale representation has been demonstrated useful for many detection tasks. Many previous detectors make use of single-scale representations, such as R-CNN [9], Fast R-CNN [11], Faster R-CNN [4], and YOLO [13]. They predict confidence and localization from the features extracted by the top-most layer within a CNN, which increases the heavy burden of the last layer. In contrast, SSD [5] uses multi-scale representations that detect objects with different scales and aspect ratios from multiple layers. For small object detection, SSD uses the features from the shallower layers, and exploits the features from the deeper layers for bigger objects detection. YOLOv3 borrows multi-scale detection from SSD and performs better in small objects.

**Contextual information**: Many previous studies have proved that contextual information plays an important role in object detection task, especially for small objects. The common method for introducing contextual information is to exploit the combined feature maps within a CNN for prediction. For example, ION [15] extracts VGG16 features from multiple layers of each region proposal using ROI pooling, and concatenates them as a fixed-size descriptor for final prediction. HyperNet [16] also adopts a similar way that use the combined feature descriptor of each region proposal for object detection. In order to further improve the accuracy of SSD, especially for small objects, some studies modify the original SSD architecture. DSSD [17] adds extra deconvolution layers at the end of SSD and the contextual information is injected by integrating every prediction layer and its deconvolution layer. FSSD [18] combines the multi-layer features from VGG-16 [6] and then generates a feature pyramid like the original SSD detection part. Feature-Fused SSD [19] extracts VGG16 features from conv4_3 and conv5_3 to replace the original conv4_3 layer. The combined features come from different layers have different levels of abstraction of input image.

MDSSD [20] (Multi-scale Deconvolutional Single Shot Detector) combines multiple high-level features with different scales simultaneously via deconvolution Fusion Block.

**Receptive field**: Receptive field (RF) refers to a region that a unit neuron in a certain layer of the network can see in the original input image [7]. Zhou *et al.* [21] were the first to introduce the concept of empirical receptive fields (ERF), and showed that the true size of RF is much smaller than the theoretical receptive field (TRF) through a data-driven approach. CSSD [7] studies the relation between TRF and ERF, and proposes DiCSSD which rapidly expands the TRF sizes of each prediction layer of the conventional SSD, ensuring that every feature point sees sufficiently large areas.

## 3. ARCHITECTURE DESIGN

As shown in Figure 2, our DDSSD architecture is built on the original SSD. In this section, we first review the structure of SSD and then introduce our DDSSD and Dilation and Deconvolution module.

### 3.1. Architecture

As illustrated in Fig.2, the boxes indicate the feature maps of SSD architecture. The original SSD detects objects by features from multi-scale layers directly, regarding these features in different levels as the same level. Actually, ConvNets have an excellent ability to extract pyramidal feature hierarchy, which has more semantic information from low to high levels. The disadvantages of this strategy are that SSD lacks the capability to capture both the local detailed features and global semantic features. However, the detector should merge the context

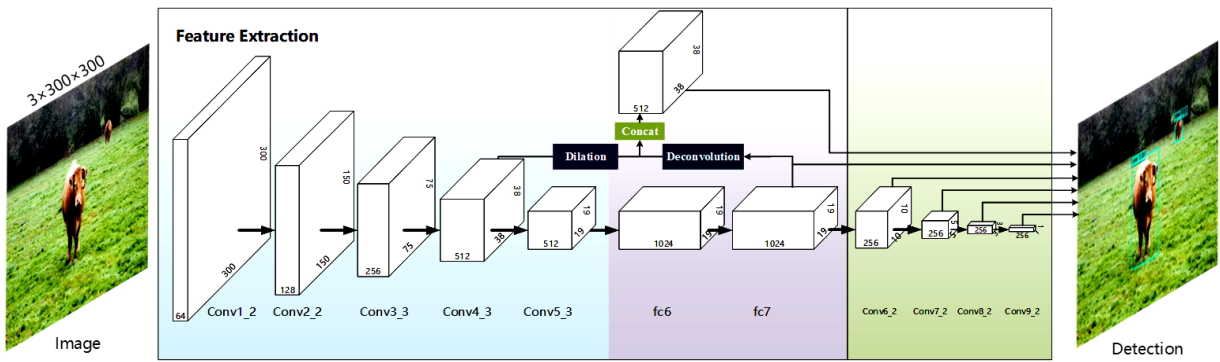

**Figure 2.** Pipeline for DDSSD.

As in SSD [5], a reduced VGG-16 [6] with two additional convolution layers (fc_6 and fc_7) is used to extract features. "Dilation" represents for Dilation Module, "Deconvolution" for Deconvolution Module and "Concat" for feature concatenation.

information and their detailed features to confirm the small objects. So, DDSSD uses the feature maps of prediction layers after fusion of context layers, rather than using them directly as in SSD.

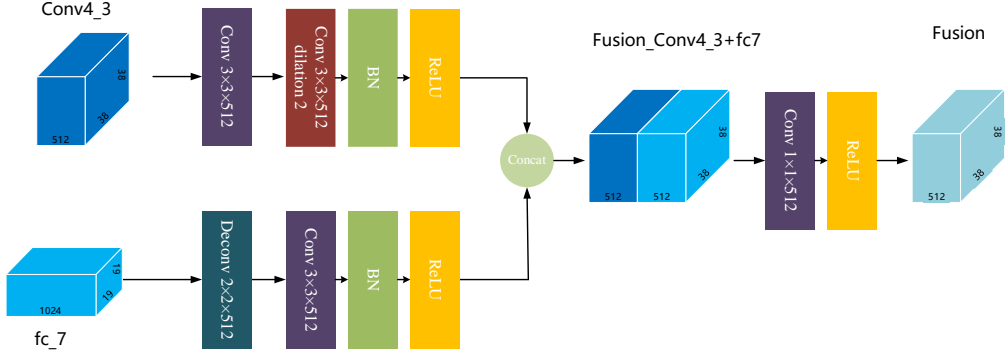

**Figure 3.** Illustration of Dilation and Deconvolution Module.
The Dilation branch including a Conv with kernel 3 × 3 and a Conv with kernel 3 × 3 and dilation 2 is utilized to enlarge the receptive field in Conv4_3. The Deconvolution module uses a Deconv with 2 × 2 kernel size and a Conv with 3 × 3 kernel to expand the resolution of feature maps from fc_7.

### 3.2. Dilation and Deconvolution Module

Instead of using the features from different scales directly, we exploit the combination of feature maps to introduce semantic information in shallow layers for detecting small objects. The pipeline of DDSSD is shown in Fig.2, and the Dilation and Deconvolution Module is illustrated in Figure 3. Deconvolution Module is utilized to expand the resolution of feature maps from deeper layer fc7, ensuring the feature maps have the same dimension which is essential to the later concatenation operation. Dilation Convolution Module is used to enlarge the receptive field of the conventional VGG [6]. After that, features from these two modules are concatenated together to replace the original feature from Conv4_3. Assuming $X_i, i \in P$ are the multi-scale features for detection, the feature fusion module can be described as follows:

$$X_f = \mathcal{C}\{\mathcal{D}(X_{Conv4\_3}), \mathcal{T}(X_{fc\_7})\} \tag{1}$$

$$loc, class = \phi_{c,l}(X_p) \quad p \in \mathcal{P} \tag{2}$$

where $\mathcal{D}$ means the Dilation Module for the features $X_{Conv4\_3}$ from layer Conv4_3. $\mathcal{T}$ is the deconvolution function, and $\mathcal{C}$ stands for the concatenation function; $\phi(c,l)$ is the method to predict object detections from the provided feature maps.

$\mathcal{D}$: Dilation convolution is often utilized in image segmentation [22, 23] to aggregate context information from different scales. According to [21, 24], the effective receptive fields (ERFs) are 2D-gaussian distributed and proved to be significantly smaller than the corresponding theoretical receptive fields (TRFs). As analyzed in CSSD [7], the ERF of Conv4_3 in SSD is only ×1.9 larger than the corresponding grid scale $\theta_p$ (not exceeding ×2.5 across all prediction layers), which motivates the need for more contexts to be integrated into the existing framework. So, we adopt a 3 × 3 convolution layer with dilation 2 to enlarge the receptive field of Conv4_3 while keeping the resolution. This module achieves a wider range of activation than the original SSD, see Fig.1.

$\mathcal{T}$: To integrate semantic information for small object detection in shallower layer, intuitively we reuse the feature maps from higher layers and merge them into the shallower ones. However, as the feature map sizes differ a lot (e.g. by a factor of 1/2 after every pooling layer with stride 2), we need to upsample maps from top layers first and ensure all sizes of feature maps are the same. Considering the convolutional part of the network as an encoding step, deconvolution or transposed convolution layer can be treated as a decoding processing step. The encode-decode structure has been shown to be particularly useful in segmentation [25]. Deconvolution has been used in many previous works like DSSD [17] and CDSSD [7]. Using deconvolution layers to integrate multi-scale information has one obvious drawback that the memory usage of network increases significantly. The reason is that the coefficients of bilinear filters and the later conv layers (which have been demonstrated to be especially useful in our experiments) take many more parameters. We additionally add one batch normalization layer that allows for more stable learning after each context layer.

## 4. EXPERIMENTAL RESULTS

### 4.1. Experimental Setup

Our DDSSD and baseline SSD are built on the PyTorch framework, and the VGG16 architecture. Our training strategies mostly follow SSD, including data augmentation, hard negative mining, scale and aspect ratios for default boxes, and loss functions (e.g., smooth L1 loss for localization and softmax loss for classification), while we slightly change our learning rate scheduling for better accommodation our DDSSD. All newly added layers are initialized with "xavier" method. The VGG16 model pre-trained on ImageNet [26] dataset for the task of image classification is reduced of fully connected layers when used as the base network of SSD. The baseline SSD is trained with the batch size of 16, and with the input size of 300×300. The training process starts with the learning rate at $10^{-3}$ for the first 80K iterations, which decreases to $10^{-4}$ and $10^{-5}$ at 100K and 120K. The momentum and weight decay are set to 0.9 and 0.0005 respectively by using SGD.

### 4.2. PASCAL VOC

We train our models on the union of PASCAL VOC2007 *trainval* and VOC2012 *trainval*, and evaluate on VOC 2007 test set. Instead of using a multi-step learning rate strategy in SSD, we follow the training strategy in [27]. We use a "warmup" strategy that gradually increases the learning rate from $10^{-6}$ to $4 \times 10^{-3}$ at the first 5 epochs. After the "warmup" phase, it goes back to the original learning rate schedule, divided by 10 at 150 and 200 epochs. The total number of training epochs is 250. We optimize the model by SGD with momentum and weight decay set to 0.9 and 0.0005 respectively.

**Table 1**. Comparison of Speed & Accuracy on PASCAL VOC2007 & 2012.

| Method | network | mAP 2007 | mAP 2012 | FPS | #Proposals | GPU | Input resolution |
|---|---|---|---|---|---|---|---|
| Faster R-CNN [4] | VGG | 73.2 | 70.4 | 7 | 6000 | Titan X | ~ 1000 × 600 |
| HyperNet [16] | VGG | 76.3 | 71.4 | 5 | - | Titan X | ~ 1000 × 600 |
| RON384++ [28] | VGG | 77.6 | 75.4 | 9 | 300 | - | ~ 1000 × 600 |
| R-FCN [12] | ResNet101 | 80.5 | 77.6 | - | - | Titan X | ~ 1000 × 600 |
| SSD [5] | VGG | 77.5 | 72.4 | 46 | 8732 | Titan X | 300 × 300 |
| DSOD [29] | DS/64-192-48-1 | 77.7 | 76.3 | 17.4 | - | Titan X | 300 × 300 |
| DiSSD300 [7] | VGG | 78.1 | - | 40.8 | 8732 | Titan X | 300 × 300 |
| MDSSD [20] | ResNet101 | 78.6 | -76.3 | 38.5 | 8732 | GTX 1080Ti | 300 × 300 |
| DSSD [17] | VGG | 78.6 | - | 11.8 | 17080 | Titan X | 321 × 321 |
| FSSD300 [18] | VGG | 78.8 | - | 65.8 | 8732 | GTX 1080Ti | 300 × 300 |
| FFSSD300 [19] | VGG | 78.8 | - | 40 | 8732 | GTX 1080Ti | 300 × 300 |
| ESSD [30] | VGG | 79.4 | - | 25 | 8732 | Titan X | 300 × 300 |
| SSD [5] | VGG | 77.5 | 72.4 | 66 | 8732 | GTX 1080 | 300 × 300 |
| DDSSD | VGG | **79.7** | **77.7** | 41 | 8732 | GTX 1080 | 300 × 300 |

As shown in Table 1, we compare the speed and accuracy of benchmarked models on PASCAL VOC 2007 & 2012 test. Our proposed Dilation and Deconvolution Module applied on top of this improved SSD, further boost performance and take little memory consumption. The proposed DDSSD network achieves the highest mAP, while maintaining a real-time speed of 41 FPS on a single nVIDIA GTX 1080 GPU, Intel(R) Xeon(R) CPU E5-2683 v3 @ 2.00GHz. Therefore, DDSSD has been the most effective model among all models compared.

### 4.3. FLIR ADAS

The recently released FLIR ADAS [1] dataset consists of a total of 10,228 total frames (8,862 for training and 1366 for validation) and 9,219 images with bounding box annotations in MS COCO [31] style, where each image if of $640 \times 512$ resolution and is captured by a FLIR Tau2 camera. 60% of the images are collected during daytime and the remaining 40% are captured during night. For the experiments, we use the training and validation splits provided in the dataset benchmark, which contains the person (22,372 instances), car (14,013 instances) and bicycle (1,205 instances). Over 58% of these instances are belonging to small objects with area under $32 \times 32$. For this dataset, we follow the same training strategy for PASCAL VOC. As shown in Table 2, we present results on FLIR ADAS Thermal Image dataset of SSD300 [5], FFSSD [19] and our DDSSD. DDSSD exceeds the other two models in a large margin, and the visualized results of SSD300 and DDSSD are illustrated in Fig.5.

**Table 2.** Results on FLIR_ADAS Thermal Image dataset.

| Method | Avg. Precision, IoU: | | | Categories, IoU:0.5 | | | Avg. Precision, Area: | | | Avg. Recall, Area: | | |
|---|---|---|---|---|---|---|---|---|---|---|---|---|
| | 0.5:0.95 | 0.5 | 0.75 | People | Bicycles | Cars | S | M | L | S | M | L |
| SSD300 [5] | 20.3 | 45.3 | 15 | 59.3 | 39.7 | 79.8 | **10.4** | 26.2 | 44.2 | 23.8 | 37.2 | 55.7 |
| FFSSD300 [19] | 19.6 | 44.6 | 14.4 | 56.9 | 42.7 | 78.6 | 9.5 | 25.5 | 44.9 | 23.1 | 34.2 | 55 |
| DDSSD300 | **22.8** | **49.9** | **17** | **62.7** | **45.5** | 79.2 | 10 | **30.1** | **46.6** | **24** | **40.3** | **56.3** |

### 4.4. Ablation study on Pascal VOC 2007

To demonstrate the effectiveness of our proposed Dilation and Deconvolution Module, we offer an ablation study on VOC 2007 in Table 3. In the conventional structure of SSD, Conv4_3 is responsible for the detection of small objects. So, FFSSD [19] combines Conv4_3 and Conv5_3 with a convolution layer and Deconvolution Module, achieving a mAP of 78.8%. Instead of using high layer semantic information from Conv5_3, we choose features from fc_7 layer and get 79.2% with the same structure in FFSSD. Replacing the convolution layer with a dilation convolution layer, we boost the performance with 0.2%. As shown in Fig.1, our assumption is that the receptive field in Conv4_3 is not enough for the detection of small objects. So, using our Dilation and Deconvolution Module, we achieve a final mAP of 79.7%.

**Table 3.** Ablation study on VOC2007 test.

| Method | SSD | | | | DDSSD |
|---|---|---|---|---|---|
| Conv4_3 + conv | | √ | √ | | |
| Conv4_3 + dilation | | | | √ | |
| Conv4_3 + Dilation Module | | | | | √ |
| Conv5_3 + Deconv Module | | √ | | | |
| fc_7 + Deconv Module | | | √ | √ | √ |
| mAP | 77.5 | 78.8 | 79.2 | 79.4 | **79.7** |

## 4.5. MS COCO

To further indicate the effectiveness of our proposed DDSSD, we evaluate DDSSD on a more general, large-scale object detection dataset, MS COCO [31]. We use trainval35k training set and minival for validation, and we show results on test-dev which is evaluated on the official evaluation server. As presented in Table 4, we get an mmAP of 28.3% over the general metric of COCO. Especially, we achieve 48.3% under IoU over 0.5 and 10.5% for small objects, which is much better than the original SSD. DDSSD exceeds SSD, FSSD, MDSSD under mmAP and AP50 significantly, but it performs worse than DSOD and RefineDet. For small objects, DDSSD outperforms all one-stage methods except MDSSD, but it achieves the highest Recall for small objects, exceeding MDSSD by a large margin. So, our DDSSD achieves a comparable performance with many state-of-the-art approaches under MS COCO. Especially for small object detection, DDSSD performs well in both average precision and average recall.

**Table 4.** Results on COCO test-dev[*].

| Method | backbone | data | AP, IoU: | | | AP., Area: | | | AR, Area: | | |
|---|---|---|---|---|---|---|---|---|---|---|---|
| | | | mAP | AP50 | AP75 | S | M | L | S | M | L |
| Faster [4] | VGG16 | trainval | 21.9 | 42.7 | - | - | - | - | - | - | - |
| RON [28] | VGG16 | trainval | 27.4 | 49.5 | 27.1 | - | - | - | - | - | - |
| Faster+++ [4] | ResNet | trainval | 34.9 | 55.7 | - | - | - | - | - | - | - |
| R-FCN [12] | ResNet | trainval | 29.9 | 51.9 | - | 10.8 | 32.8 | 45 | - | - | - |
| SSD300 [5] | VGG16 | trainval35k | 25.1 | 43.1 | 25.8 | 6.6 | 25.9 | 41.4 | 11.2 | 40.4 | 58.5 |
| MDSSD [20] | VGG16 | trainval35k | 26.8 | 45.9 | 27.7 | 10.8 | 27.5 | 40.8 | 15.8 | 42.3 | 56.3 |
| FSSD300 [18] | VGG16 | trainval35k | 27.1 | 47.7 | 27.8 | 8.7 | 29.2 | 42.2 | 15.9 | 44.2 | 58.6 |
| SSD321 [5] | ResNet | trainval35k | 28 | 45.4 | 29.3 | 6.2 | 28.3 | 49.3 | 11.5 | 43.3 | 64.9 |
| DDSSD300 | VGG16 | trainval35k | **28.3** | **48.3** | 29.2 | 10.5 | 29.7 | 42.1 | 16.7 | 43.5 | 57.8 |

[*]'S', 'M' and 'L' stand for small, medium and large respectively, and 'mmAP', 'AP50' and 'AP75' mean average precision of IoU =0.5:0.95, IoU=0.5 and IoU=0.75 respectively. trainval35k is obtained by removing the 5k minival set from trainval.

## 4.6. Visualization

We present some detection results with scores above 0.6 on PASCAL VOC2007 test and FLIR ADAS validation with SSD300 [5] and DDSSD. Different colors of the bounding boxes indicate different object categories.

As we can see from Fig.4 and Fig.5, our DDSSD has two distinct improvements based on original SSD. The first is the improvement of the detection of small targets, mainly because our feature fusion strategy provides rich semantic information for the shallow layers, which helps a lot for detecting small objects. The second is that the DDSSD has a higher score for the object which can be detected by both models, mainly because the Dilation Module enlarges the receptive field which learns the relations between objects within one image.

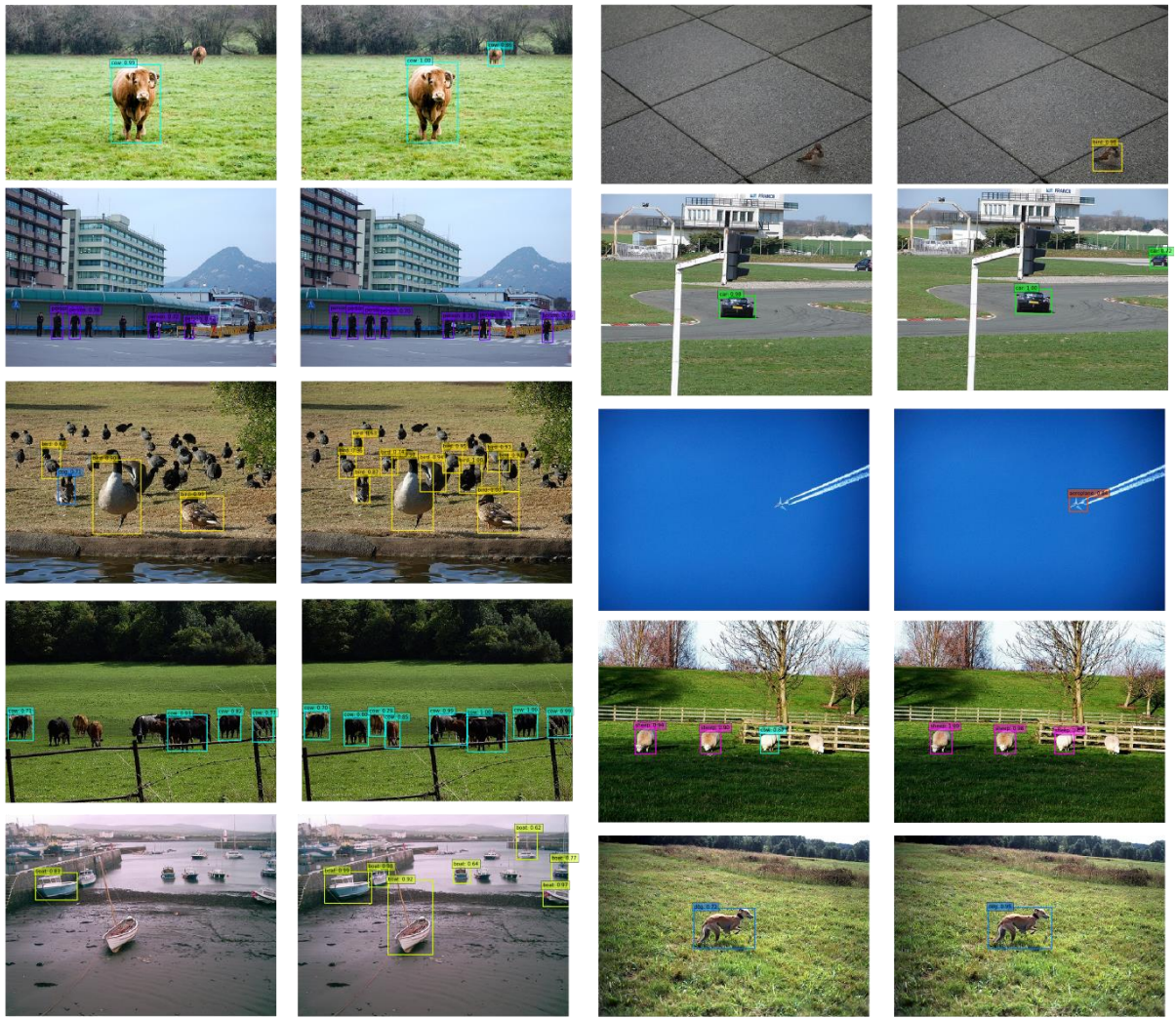

**Figure 4**. Visualized examples in PASCAL VOC 2007 test with confidence above 0.6. For each group, the left side is the result of SSD300 and right side is the result of our DDSSD.

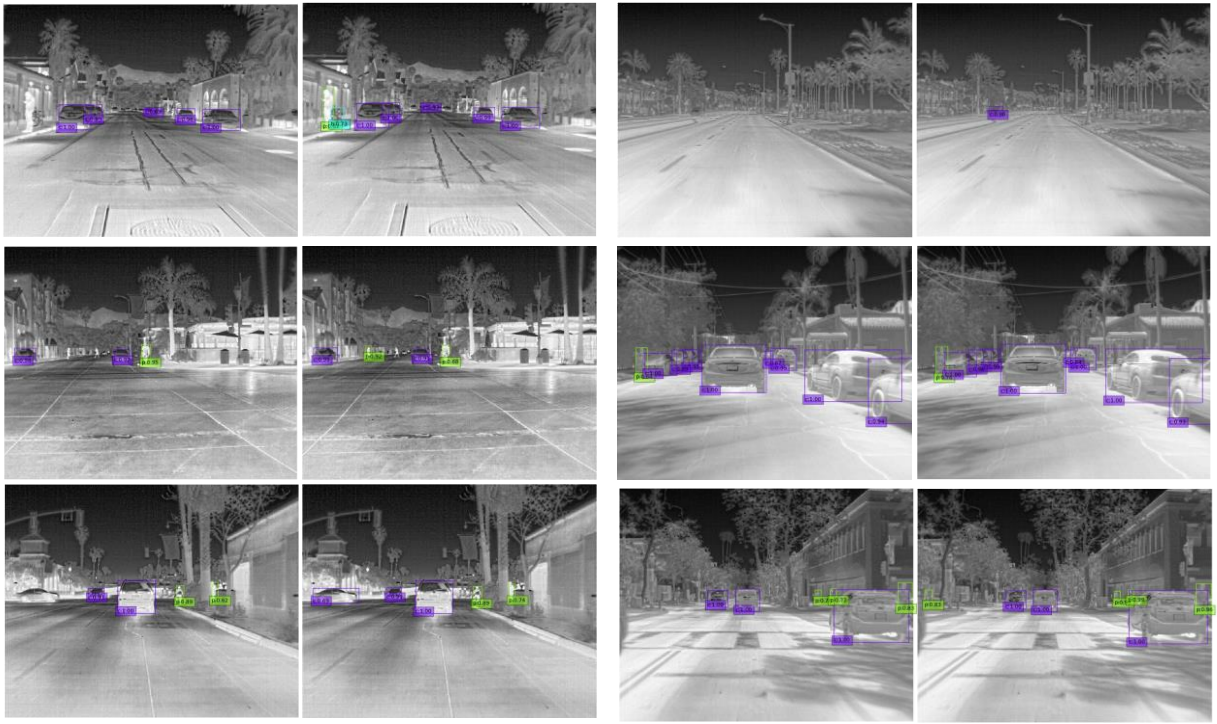

**Figure 5.** Visualized examples in FLIR ADAS validation with confidence above 0.6,
where: 'p', 'b', and 'c' represent for 'person', 'bicycle', and 'car' respectively.
For each group, the left side is the result of SSD300 [5] and right side is DDSSD.

## 4. CONCLUSION AND FUTURE WORK

In this paper, we aim to design a fast and efficient detector for small objects. First, we investigate the framework to fuse different features together, and then we propose a novel and simple feature fusion strategy to introduce semantic information into the shallow layer of SSD. Specifically, Dilation Module is responsible for expand the receptive field and Deconvolution Module is designed to introduce semantic information from higher layer. Experimental results demonstrate that our DDSSD improves conventional SSD a lot and outperforms several other state-of-the-art object detectors both in accuracy and efficiency.

Our future work will focus on the improve the inference speed of DDSSD through model compression with techniques like pruning, quantization and knowledge distillation. These methods have been demonstrated useful in classification network, accelerating the model significantly with little accuracy drop. However, they have not been tested in an object detection network.

## FUNDING

This work was partly supported by Innovation Fund for Graduate of Nanchang University under Grant CX2018145.

## CONFLICT OF INTEREST

The authors declare that they have no conflict of interest.